\documentclass[journal,transmag]{IEEEtran}

\usepackage{graphicx}
\ifCLASSOPTIONcompsoc
  \usepackage[caption=false,font=normalsize,labelfont=sf,textfont=sf]{subfig}
\else
  \usepackage[caption=false,font=footnotesize]{subfig}
\fi

\begin{document}

\title{A Novel Neural Network Model Specified for Representing Logical Relations}

\author{\IEEEauthorblockN{Gang Wang}
\thanks{Corresponding author: G. Wang (email: f\_lag@buaa.edu.cn).}}

\markboth{}%
{}

\IEEEtitleabstractindextext{%
\begin{abstract}
With computers to handle more and more complicated things in variable environments, it becomes an urgent requirement that the artificial intelligence has the ability of automatic judging and deciding according to numerous specific conditions so as to deal with the complicated and variable cases. ANNs inspired by brain is a good candidate. However, most of current numeric ANNs are not good at representing logical relations because these models still try to represent logical relations in the form of ratio based on functional approximation. On the other hand, researchers have been trying to design novel neural network models to make neural network model represent logical relations. In this work, a novel neural network model specified for representing logical relations is proposed and applied. New neurons and multiple kinds of links are defined. Inhibitory links are introduced besides exciting links. Different from current numeric ANNs, one end of an inhibitory link connects an exciting link rather than a neuron. Inhibitory links inhibit the connected exciting links conditionally to make this neural network model represent logical relations correctly. This model can simulate the operations of Boolean logic gates, and construct complex logical relations with the advantages of simpler neural network structures than recent works in this area. This work provides some ideas to make neural networks represent logical relations more directly and efficiently, and the model could be used as the complement to current numeric ANN to deal with logical issues and expand the application areas of ANN.
\end{abstract}

}

\maketitle

\IEEEdisplaynontitleabstractindextext

\IEEEpeerreviewmaketitle

\section{Introduction}

With computers to handle more and more complicated things in variable environments like driverless car and advanced medical diagnosis expert system, higher artificial intelligence becomes an urgent requirement for industries and a hot research point in academical area of computers. To solve the above complicated issues, they wish computers have the ability of automatic judging and deciding according to numerous specific conditions. In reality, biological brain is a natural advanced intelligent system which can make logical judge and decision according to specific conditions, thus it is very helpful for promoting the artificial intelligence by mimicking biological brain. Therefore, research on brain-inspired intelligence has been carried out as a part of various brain projects launched by multiple countries and areas like European HBP (Human Brain Project) \cite{Markram, HBP}, American BRAIN (Brain Research through Advancing Innovative Neurotechnologies) \cite{Bargmann} and MIcrONS (Machine Intelligence from Cortical Networks) \cite{IARPA} and Chinese Brain project \cite{Poo}. International IT companies like Google, Facebook and Baidu have also launched their own projects on brain-inspired intelligence.

Logical representation and reasoning are the important abilities of biological brain to let human make judge according to specific conditions embodied in \emph{if-then} rules. However, it is still difficult and not efficient for most of current ANN (Artificial Neural Network) models based on functional approximation to deal with logical representations used in knowledge representation \cite{Garcez1, Francois, Chenghao}. In this paper, we refer to them as numeric ANN. These ANN models are good at representing numeric relations which are used to describe the ratio between things, and have been successfully applied in perceiving intelligence like image recognition \cite{Sun, Parkhi, He}and speech recognition \cite{Mohamed, Gehring}. On the other hand, these ANN models are not good at representing logical relations which are used to describe the sequence of things solved by another AI branch symbolic logic as cognitive intelligence \cite{Luger, Michael}. The reason is the functional-approximation based ANN models, aimed for representing numeric relations, are not suitable for representing and storing logical relations in their neural network structures according to the "no free lunch" theorem \cite{Wolpert1, Wolpert2}. Aiming to make the neural network have the ability to deal with logical issues, researchers have been trying to design novel neural network models from two aspects: neural components and neural connecting styles to make neural network model can represent and store logical relations \cite{Besold, Besold1}. Through the research, they want to combine symbolic logic with ANN which are yet developing individually, and they want to apply ANN into more application areas like knowledge representation and reasoning, expert systems, semantic web and cognitive modelling and robotics. Several approaches have been proposed [20]-[27]. 
Most recently, the advances in designing logic gates by simple neurons has been proposed in \cite{Tao}, the neural network model is based on SN P systems (Spiking Neural P systems) \cite{Ionescu,Ionescu1}, and emulates the operations of Boolean logic gates with astrocyte-like control. However, the initial purpose of the fundamental computational system, from which this NN model comes, is not for representing  logical relations. As a consequence, this NN model deriving from the system is complex and undirect to represent logical relations. Every logical components to emulate basic logical operations of the model are heave-weight with excessive neurons and links, not to mention the heavier neural network structures constructed by these components to represent more complex logical expressions. Therefore, we apply PLDNN (Probabilistic Logical Dynamical Neural Network), a ANN model specified for representing logical relations directly, to emulates the operations of Boolean logic gates. In the model of PLDNN, new neurons and links are defined specified for representing logical relations. Inhibitory links are introduced besides exciting links. Different from current numeric ANNs, one end of an inhibitory link connects an exciting link rather than a neuron. Inhibitory links inhibit the connected exciting links conditionally to make PLDNN represent logical relations correctly. Compared with SN P systems with astrocytes-like control, simpler neural network structure of PLDNN are formed to represent the same logical expressions without less neurons and links than the model. The intention of this work is to provide some ideas to make neural networks represent logical relations more directly and efficiently, and PLDNN could be used as the complement to current numeric ANN to deal with logical issues and expand the application areas of ANN. This work only refers to the logical features of PLDNN to represent logical relations. The other features of PLDNN can be seen in \cite{Wang} if interested, including the probabilistic feature to deal with uncertainty and the dynamical feature for automatic network construction.

The remainder of this paper is organized as follows. Section 2 introduces how to design PLDNN specified for representing logical relations directly, including its neurons representing things and links representing relations between things. Section 3 argues how PLDNN are constructed to emulate operations of logic gates, and makes comparison with SN P systems with astrocytes-like control to show the simplicity of PLDNN in representing logical relations. Section 4 shows the simpler neural network structure of PLDNN than SN P systems with astrocytes-like control in representing more complex logical expressions. Finally, Section 5 concludes.
\section{Components in PLDNN specified for representing logical relations}

\begin{figure}[!t]
\begin{center}
\includegraphics[width=0.9\columnwidth]{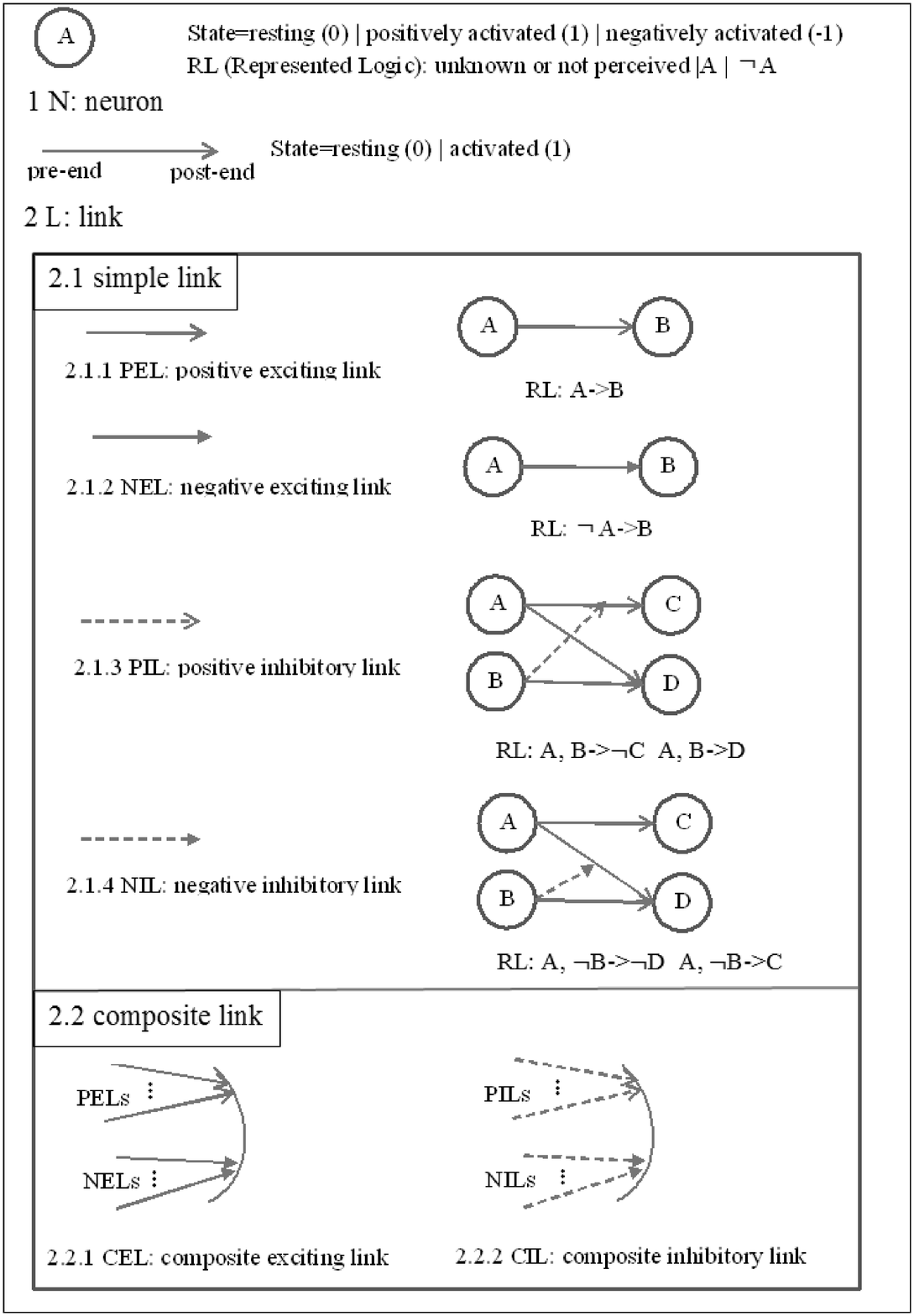}
\caption{Basic components  of  PLDNN to represent logical relations}
\label{Basic components}
\end{center}
\end{figure}

PLDNN are firstly introduced before discussing how to emulate operations of logic gates by it. The components in PLDNN are shown in Figure~\ref{Basic components}. As a subtype of ANN, PLDNN also has two super types: 1) neurons representing the things and 2) links connecting neurons to represent relations between things. In order to represent logical relations, PLDNN are designed to have multiple kinds of links, including exciting and inhibitory links specified for representing logical relations between things. Different from the link in current numeric ANN, the pre-end of IL (Inhibitory link) connects the neuron, and its post-end connects EL (Exciting link). This connection style of IL can inhibit EL connected by it from exciting EL's post-end neuron so as to make PLDNN represent the logical relations correctly. Therefore, from the design aspect, PLDNN has instinctive advantages on representing logical relations using its specified components.

The neuron $\sigma$ is used for representing a thing defined by users, it has three states: resting, positively activated and negatively activated, indicated by 0, 1 and -1. In general, the state of the neuron is in the resting state. The word 'resting' refers to the term in the biological neuron network. When a thing A happens and perceived by PLDNN, the state of the neuron A becomes positively activated achieving the goal that PLDNN represents the logic A. Similarly, when a thing A doesn't happen and perceived by PLDNN, the state of the neuron A becomes negatively activated achieving the goal that PLDNN represents the logic $\neg A$.

There are two types of links in PLDNN: exciting and inhibitory link. Unlike the way that the links in current numeric ANNs have no state,
the link in PLDNN also has states. The link has two states: resting and activated, indicated by 0 and 1. When the link is in the activated state, it can make effects on its post-end. The triggers of these links turning into the activated state are different as follows:
\begin{itemize}
\item PEL.state =1 when its pre-end neuron.state= 1
\item NEL.state =1 when its pre-end neuron.state= -1
\item PIL.state =1 when its pre-end neuron.state= 1
\item NIL.state =1 when its pre-end neuron.state= -1
\end{itemize}

These triggers are defined according to the requirements of representing logical relations. We take PEL to illustrate. For a PEL, when its pre-end neuron.state=1, the PEL is activated and makes effects on its post-end neuron to represent the logical relation $A\rightarrow B$ shown in Figure~\ref{Basic components}. When the pre-end neuron.state= 0, the PEL is not activated and makes no effects on its post-end neurons.

Multiple simple ELs can be combined together to form composite ELs to fulfill complex excitement. In the similar way, multiple simple ILs can be combined together to form composite ILs to fulfill complex inhibition. Then the interaction of the two composite links can represent complex logical relations. The triggers of composite links turning into the activated state are as follows:
\begin{itemize}
\item CEL.state =1 when the states of all simple ELs contained in this CEL are 1
\item CIL.state =1 when the states of all simple ILs contained in this CIL are 1
\end{itemize}

An example of to representing mutliple logical relations by using the components of PLDNN is shown in Figure~\ref{components  of  PLDNN}. When the things A and B happen, the neuron A and B are positively activated, PELs and PILs whose pre-end neurons are A or B are activated next. Then the neurons A and B will excite the directed linked neurons. The neuron A will excite D, E by $PEL_{AD}$ and $PEL_{AE}$. The neuron B will excite D by the $PEL_{BD}$. To let PLDNN to represent the logical relation $A, B\rightarrow D$, the neuron B prevents A from activating E by $PIL_{B,PEL_{AE}}$. The inhibitory mechanism makes PLDNN represent the right logical relations by the interactions between neurons through multiple kinds of links.
\begin{figure}[!h]
\begin{center}
\includegraphics[width=0.9\columnwidth]{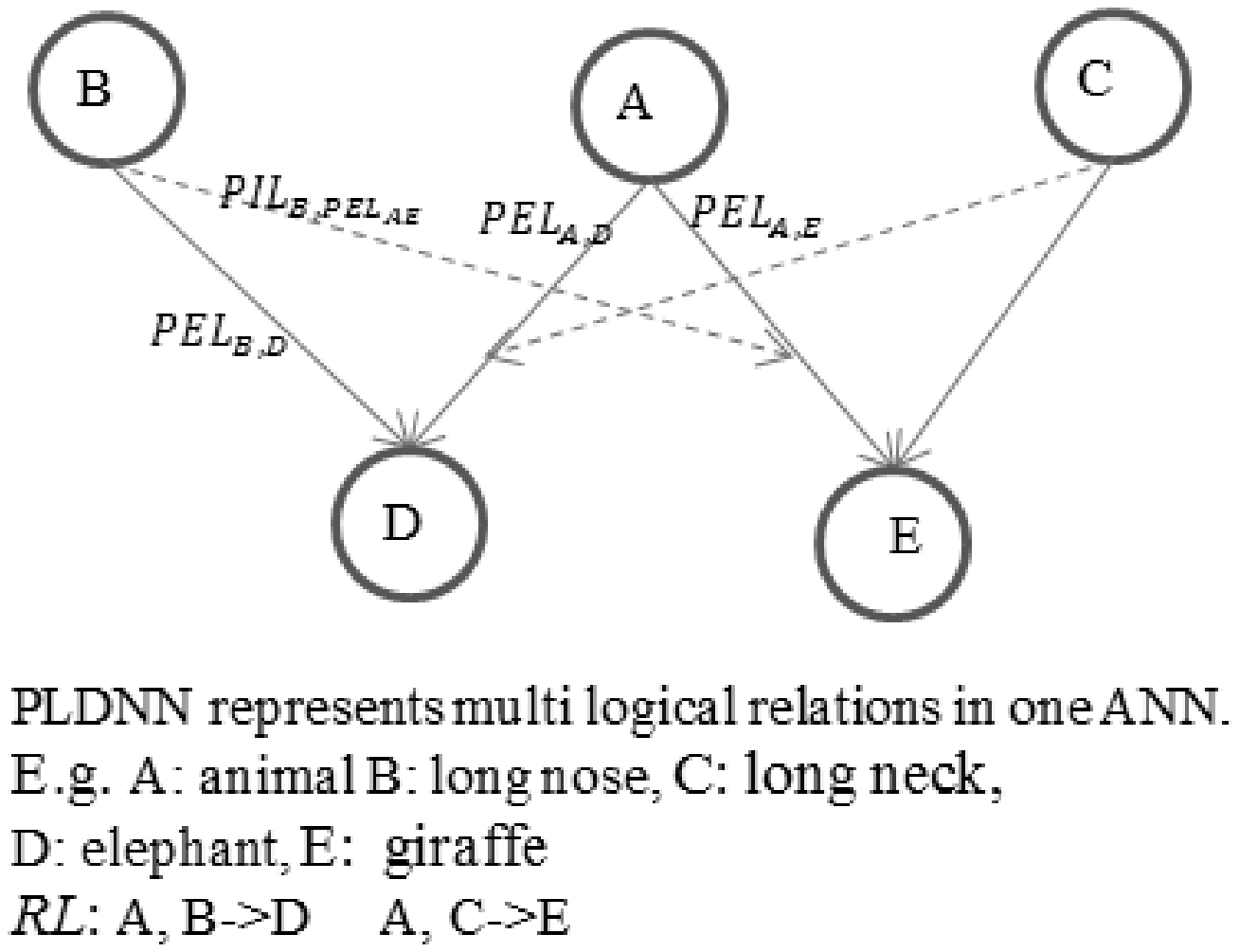}
\caption{Using components  of  PLDNN to represent logical relations}
\label{components of PLDNN}
\end{center}
\end{figure}
\section{Simulating logic gates}
\label{Simulating logic gates}
In this section, it discusses how PLDNN are constructed to emulate operations of logic gates AND, OR, NOT, NOR, XOR and NAND, respectively.

\begin{figure*}[!t]
\centering
\subfloat[PLDNN $\prod_{AND}$]{\includegraphics[width=0.9\columnwidth]{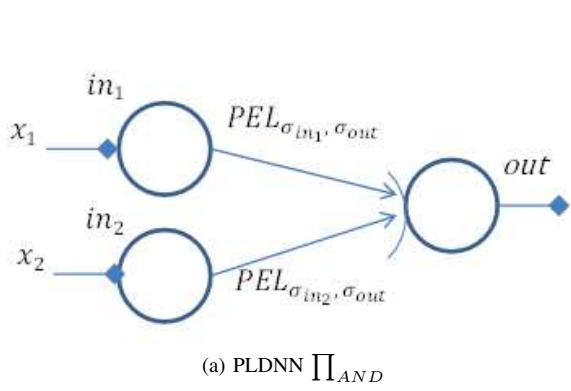}%
\label{PLDNN AND}}
\hfil
\subfloat[SN P system with astrocyte-like control $\prod_{AND}$\cite{Tao}]{\includegraphics[width=0.9\columnwidth]{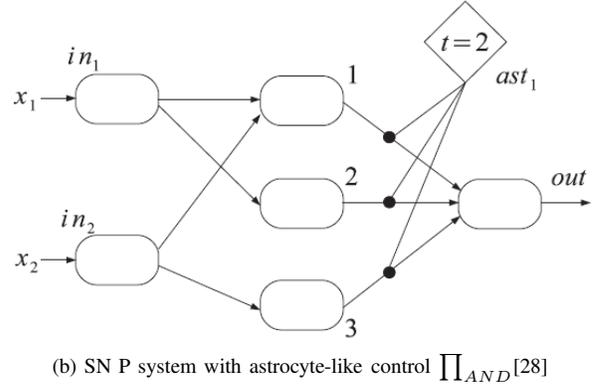}%
\label{SN P system with astrocyte-like control AND}}
\caption{Simulating a logic AND gate.}
\label{Simulating a logic AND gate.}
\end{figure*}

\begin{flushleft}
  \textbf{Theorem 3.1.} The logic AND gate can be emulated by a simple PLDNN having three neurons and one CEL with two PELs.
\end{flushleft}
\begin{flushleft}
\textbf{Proof.} An PLDNN $\prod_{AND}$ is constructed to emulate the logic gate AND, shown in Figure~\ref{PLDNN AND}. The system $\prod_{AND}$ has two input neurons $\sigma_{in1}$ and $\sigma_{in2}$, and one output neuron $\sigma_{out}$. In the following, all the four cases of inputs to logic ADD gate are considered.
\end{flushleft}
\begin{itemize}
  \item If the inputs are $x_{1}$ = -1 , $x_{2}$ = -1 , the neurons $\sigma_{in_{1}}$ and $\sigma_{in_{2}}$ representing them are negatively activated, i.e. $\sigma_{in_{1}}$.state=-1 and $\sigma_{in_{2}}$.state=-1. The link $PEL_{\sigma_{in_{1}}, \sigma_{out}}$ of $\sigma_{in_{1}}$ is not activated, i.e. $PEL_{\sigma_{in_{1}}, \sigma_{out}}$.state is 0 according to the activating condition of PEL. So does the $PEL_{\sigma_{in_{2}}, \sigma_{out}}$, and $PEL_{\sigma_{in_{2}}, \sigma_{out}}$.state is 0. At this moment, the composting exciting link $CEL_{\{\sigma_{in_{1}}, \sigma_{in_{2}}\}, \sigma_{out}}$ is not activated according to the activating condition of CEL that all simple ELs should be in the activated state. Consequently, the output neuron $\sigma_{out}$, the post-end of $CEL_{\{\sigma_{in_{1}}, \sigma_{in_{2}}\}, \sigma_{out}}$, is not activated indicating that computation result of the logic AND gate is FALSE.
  \item If the inputs are $x_{1}$ = -1 , $x_{2}$ = 1 , the neuron $\sigma_{in_{1}}$ representing $x_{1}$ is negatively activated and the neuron $\sigma_{in_{2}}$ representing $x_{2}$ is positively activated , i.e. $\sigma_{in_{1}}$.state=-1 and $\sigma_{in_{2}}$.state=1. The link $PEL_{\sigma_{in_{1}}, \sigma_{out}}$ of $\sigma_{in_{1}}$ is not activated, i.e. $PEL_{\sigma_{in_{1}}, \sigma_{out}}$.state is 0. $PEL_{\sigma_{in_{2}}, \sigma_{out}}$ is activated, and $PEL_{\sigma_{in_{2}}, \sigma_{out}}$.state is 1. At this moment, the composting exciting link $CEL_{\{\sigma_{in_{1}}, \sigma_{in_{2}}\}, \sigma_{out}}$ is not activated according to the activating condition of CEL that all simple ELs should be in the activated state. Consequently, the output neuron $\sigma_{out}$, the post-end of $CEL_{\{\sigma_{in_{1}}, \sigma_{in_{2}}\}, \sigma_{out}}$, is not activated indicating that computation result of the logic AND gate is FALSE.
  \item If the inputs are $x_{1}$ = 1 , $x_{2}$ = -1 , the neuron $\sigma_{in_{1}}$ representing $x_{1}$ is positively activated and the neuron $\sigma_{in_{2}}$ representing $x_{2}$ is negatively activated , i.e. $\sigma_{in_{1}}$.state=1 and $\sigma_{in_{2}}$.state=-1. The computation process of system  AND is quite similar to the case of inputs being $x_{1}$ = -1 , $x_{2}$ = 1, just exchanging their positions. In this case, the output neuron $\sigma_{out}$, the post-end of $CEL_{\{\sigma_{in_{1}}, \sigma_{in_{2}}\}, \sigma_{out}}$, is not activated indicating that computation result of the logic AND gate is FALSE.
  \item If the inputs are $x_{1}$ = 1 , $x_{2}$ = 1 , the neurons $\sigma_{in_{1}}$ and $\sigma_{in_{2}}$ representing them are positively activated, i.e. $\sigma_{in_{1}}$.state=1 and $\sigma_{in_{2}}$.state=1. The link $PEL_{\sigma_{in_{1}}, \sigma_{out}}$ of $\sigma_{in_{1}}$ is activated, i.e. $PEL_{\sigma_{in_{1}}, \sigma_{out}}$.state is 1 according to the activating condition of PEL. So does the $PEL_{\sigma_{in_{2}}, \sigma_{out}}$, and $PEL_{\sigma_{in_{2}}, \sigma_{out}}$.state is 1. At this moment, the composting exciting link $CEL_{\{\sigma_{in_{1}}, \sigma_{in_{2}}\}, \sigma_{out}}$ is activated according to the activating condition of CEL that all simple ELs should be in the activated state. Eventually, the output neuron $\sigma_{out}$, the post-end of $CEL_{\{\sigma_{in_{1}}, \sigma_{in_{2}}\}, \sigma_{out}}$, is activated indicating that computation result of the logic AND gate is TRUE.
\end{itemize}

Based on the description of the PLDNN $\prod_{AND}$ above, it is clear that the PLDNN $\prod_{AND}$ can correctly emulate the operation of a logic AND gate, and comparing Figure~\ref{PLDNN AND} with Figure~\ref{SN P system with astrocyte-like control AND}, the neural network structure of PLDNN is simpler than SN P systems with astrocyte-like control in emulating the logic gate AND which has six neurons and more links.

\begin{figure*}[!t]
\centering
\subfloat[PLDNN $\prod_{OR}$]{\includegraphics[width=0.9\columnwidth]{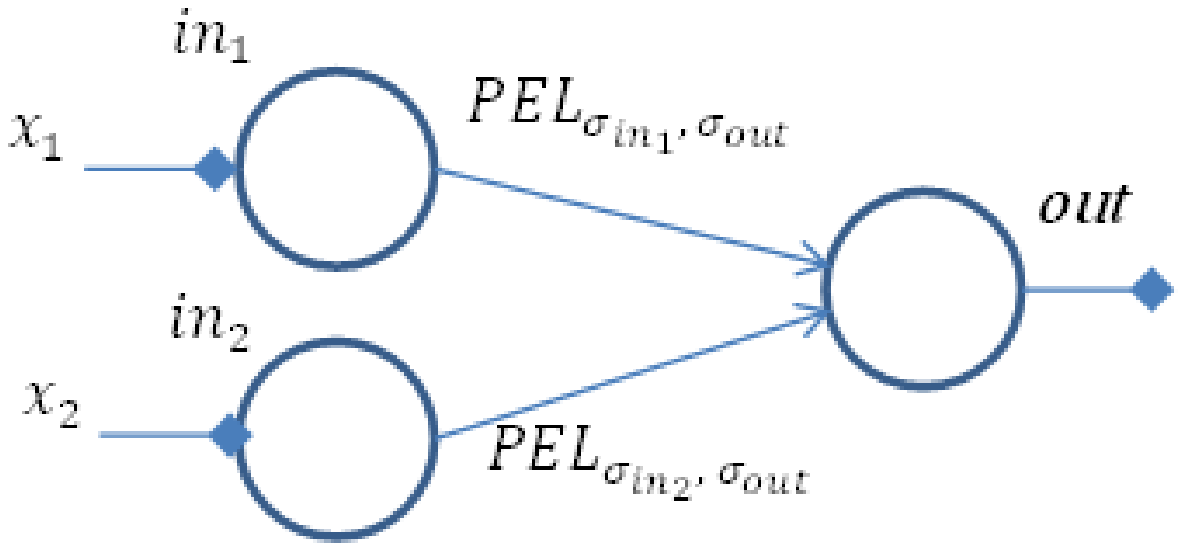}%
\label{PLDNN OR}}
\hfil
\subfloat[SN P system with astrocyte-like control $\prod_{OR}$\cite{Tao}]{\includegraphics[width=0.9\columnwidth]{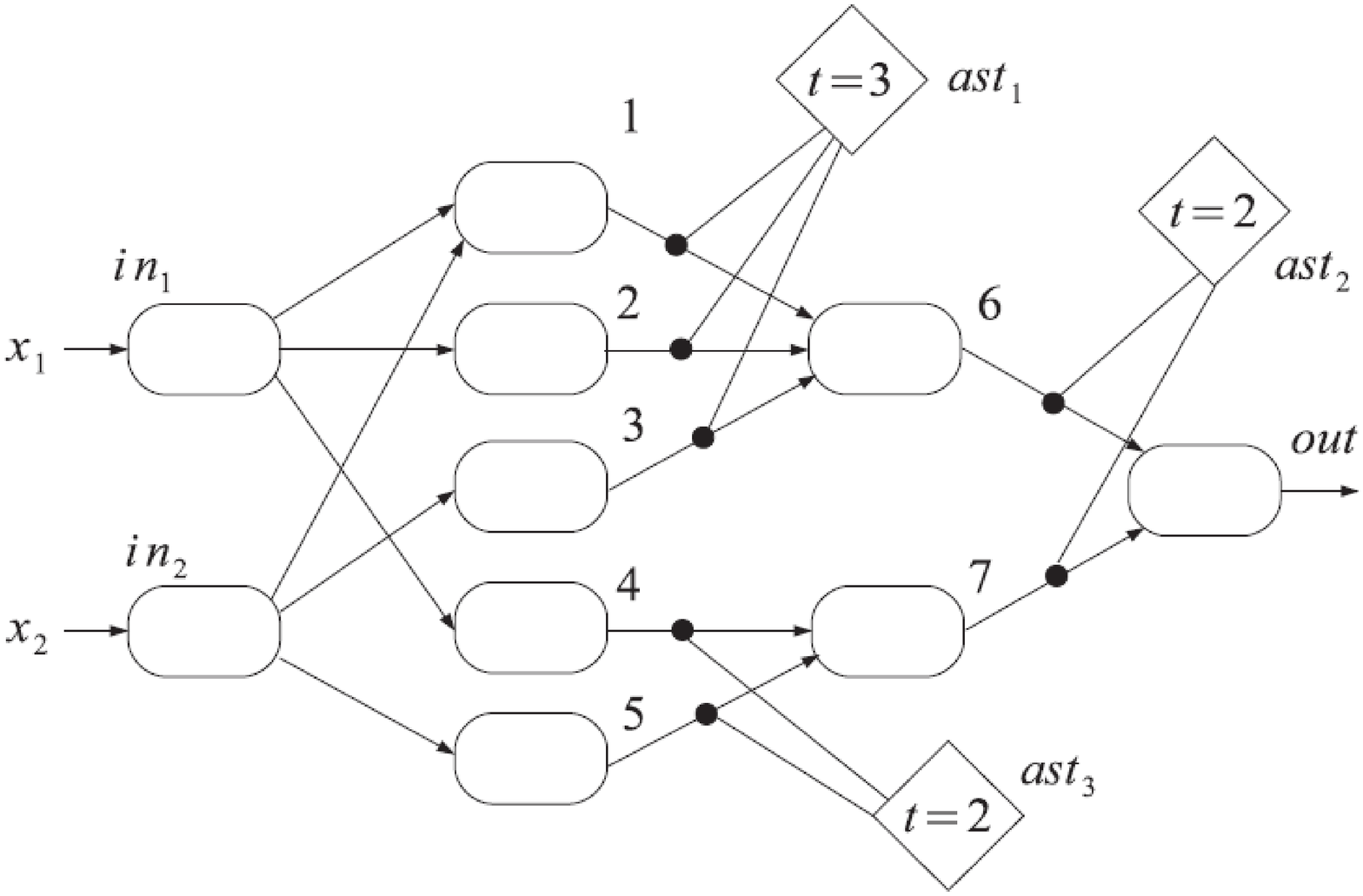}%
\label{SN P system with astrocyte-like control OR}}
\caption{Simulating a logic OR gate.}
\label{Simulating a logic OR gate.}
\end{figure*}

This concludes the proof.
\begin{flushleft}
  \textbf{Theorem 3.2.} The logic OR gate can be emulated by a simple PLDNN having three neurons and two PELs.
\end{flushleft}
\begin{flushleft}
\textbf{Proof.} An PLDNN $\prod_{OR}$ is constructed to emulate the logic gate OR, shown in Figure~\ref{PLDNN OR}. The system $\prod_{OR}$ has two input neurons $\sigma_{in1}$ and $\sigma_{in2}$, and one output neuron $\sigma_{out}$. In the following, all the four cases of inputs to logic OR gate are considered.
\end{flushleft}
\begin{itemize}
  \item If the inputs are $x_{1}$ = -1 , $x_{2}$ = -1 , the neurons $\sigma_{in_{1}}$ and $\sigma_{in_{2}}$ representing them are negatively activated, i.e. $\sigma_{in_{1}}$.state=-1 and $\sigma_{in_{2}}$.state=-1. The link $PEL_{\sigma_{in_{1}}, \sigma_{out}}$ of $\sigma_{in_{1}}$ is not activated, i.e. $PEL_{\sigma_{in_{1}}, \sigma_{out}}$.state is 0 according to the activating condition of PEL. So does the $PEL_{\sigma_{in_{2}}, \sigma_{out}}$, and $PEL_{\sigma_{in_{2}}, \sigma_{out}}$.state is 0. Consequently, the output neuron $\sigma_{out}$, the post-end of $PEL_{\sigma_{in_{1}}, \sigma_{out}}$ and $PEL_{\sigma_{in_{2}}, \sigma_{out}}$, is not activated indicating that computation result of the logic OR gate is FALSE.
  \item If the inputs are $x_{1}$ = -1 , $x_{2}$ = 1 , the neuron $\sigma_{in_{1}}$ representing $x_{1}$ is negatively activated and the neuron $\sigma_{in_{2}}$ representing $x_{2}$ is positively activated , i.e. $\sigma_{in_{1}}$.state=-1 and $\sigma_{in_{2}}$.state=1. The link $PEL_{\sigma_{in_{1}}, \sigma_{out}}$ of $\sigma_{in_{1}}$ is not activated, i.e. $PEL_{\sigma_{in_{1}}, \sigma_{out}}$.state is 0. $PEL_{\sigma_{in_{2}}, \sigma_{out}}$ is activated, and $PEL_{\sigma_{in_{2}}, \sigma_{out}}$.state is 1. Consequently, the output neuron $\sigma_{out}$ is activated by $PEL_{\sigma_{in_{2}}, \sigma_{out}}$, indicating that computation result of the logic OR gate is TRUE.
  \item If the inputs are $x_{1}$ = 1 , $x_{2}$ = -1 , the neuron $\sigma_{in_{1}}$ representing $x_{1}$ is positively activated and the neuron $\sigma_{in_{2}}$ representing $x_{2}$ is negatively activated , i.e. $\sigma_{in_{1}}$.state=1 and $\sigma_{in_{2}}$.state=-1. The computation process of system  OR is quite similar to the case of inputs being $x_{1}$ = -1 , $x_{2}$ = 1, just exchanging their positions. In this case, the output neuron $\sigma_{out}$ is activated by $PEL_{\sigma_{in_{1}}, \sigma_{out}}$ indicating that computation result of the logic OR gate is TRUE.
  \item If the inputs are $x_{1}$ = 1 , $x_{2}$ = 1 , the neurons $\sigma_{in_{1}}$ and $\sigma_{in_{2}}$ representing them are positively activated, i.e. $\sigma_{in_{1}}$.state=1 and $\sigma_{in_{2}}$.state=1. The link $PEL_{\sigma_{in_{1}}, \sigma_{out}}$ of $\sigma_{in_{1}}$ is activated, i.e. $PEL_{\sigma_{in_{1}}, \sigma_{out}}$.state is 1 according to the activating condition of PEL. So does the $PEL_{\sigma_{in_{2}}, \sigma_{out}}$, and $PEL_{\sigma_{in_{2}}, \sigma_{out}}$.state is 1. Eventually, the output neuron $\sigma_{out}$ is activated by $PEL_{\sigma_{in_{1}}, \sigma_{out}}$ and $PEL_{\sigma_{in_{2}}, \sigma_{out}}$ indicating that computation result of the logic OR gate is TRUE.
\end{itemize}

Based on the description of the PLDNN $\prod_{OR}$ above, it is clear that the PLDNN $\prod_{OR}$ can correctly emulate the operation of a logic OR gate, and comparing Figure~\ref{PLDNN OR} with Figure~\ref{SN P system with astrocyte-like control OR}, the neural network structure of PLDNN is simpler than SN P systems with astrocyte-like control in emulating the logic gate OR which has ten neurons and more links.

\begin{figure*}[!t]
\centering
\subfloat[PLDNN $\prod_{NOT}$]{\includegraphics[width=0.9\columnwidth]{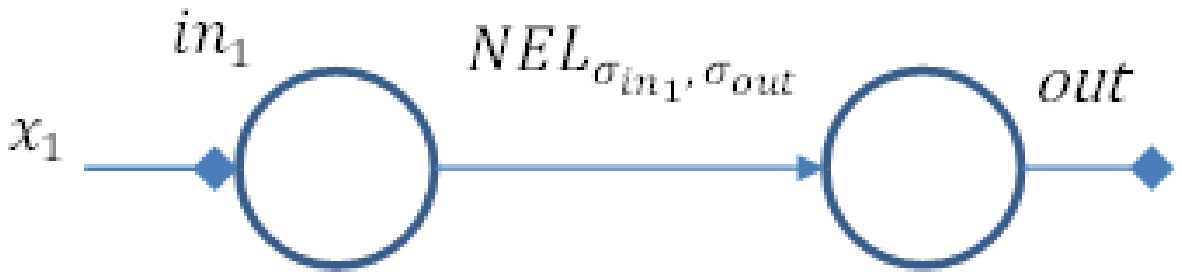}%
\label{PLDNN NOT}}
\hfil
\subfloat[SN P system with astrocyte-like control $\prod_{NOT}$\cite{Tao}]{\includegraphics[width=0.9\columnwidth]{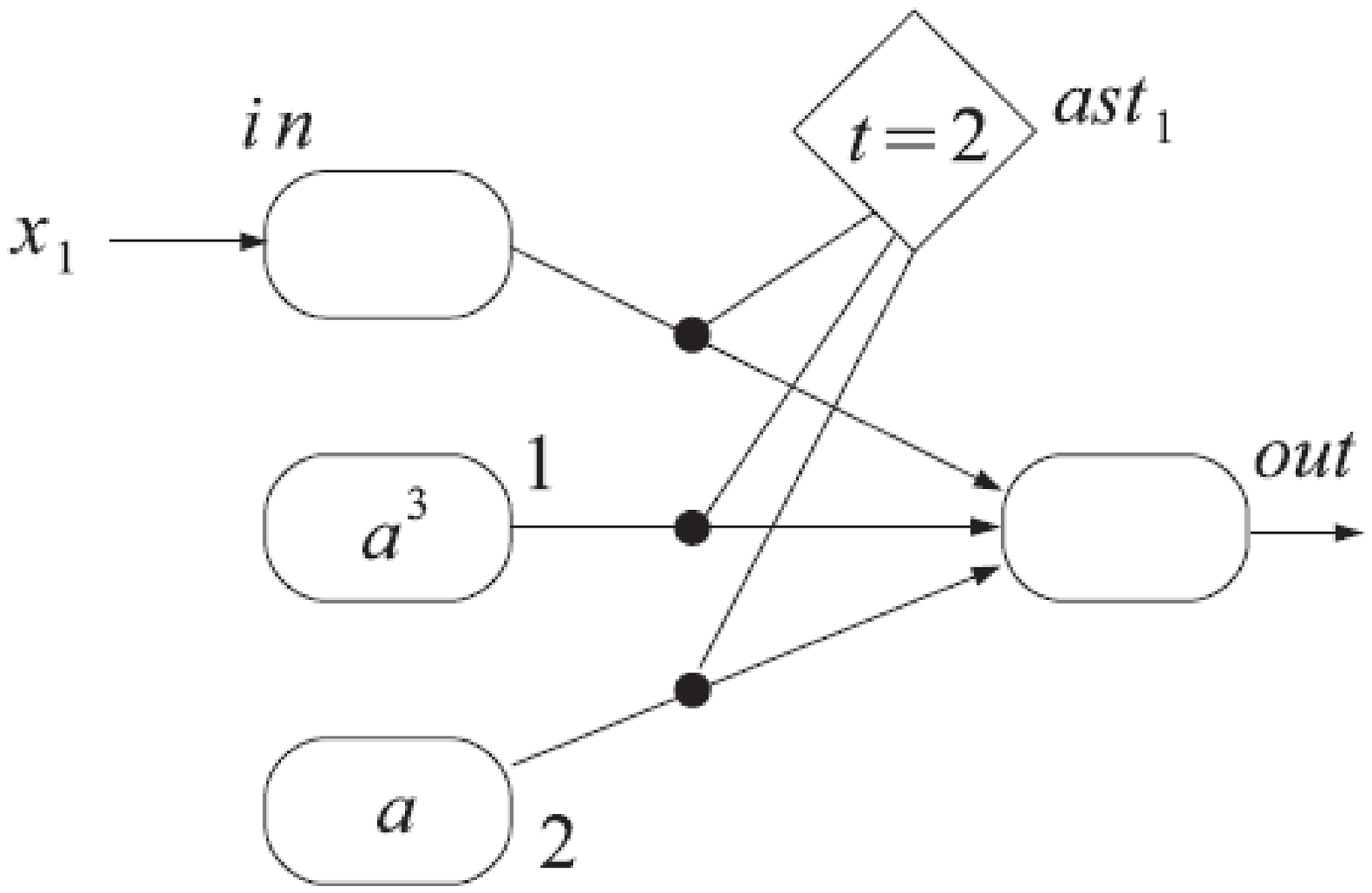}%
\label{SN P system with astrocyte-like control NOT}}
\caption{Simulating a logic NOT gate.}
\label{Simulating a logic NOT gate.}
\end{figure*}

This concludes the proof.
\begin{flushleft}
  \textbf{Theorem 3.3.} The logic NOT gate can be emulated by a simple PLDNN having two neurons and one NEL.
\end{flushleft}
\begin{flushleft}
\textbf{Proof.} An PLDNN $\prod_{NOT}$ is constructed to emulate the logic gate NOT, shown in Figure~\ref{PLDNN NOT}. The system $\prod_{NOT}$ has one input neuron $\sigma_{in1}$, and one output neuron $\sigma_{out}$. In the following, all the two cases of the input to logic NOT gate are considered.
\end{flushleft}
\begin{itemize}
  \item If the inputs are $x_{1}$ = -1, the neurons $\sigma_{in_{1}}$ representing $x_{1}$ is negatively activated, i.e. $\sigma_{in_{1}}$.state=-1. The link $NEL_{\sigma_{in_{1}}, \sigma_{out}}$ of $\sigma_{in_{1}}$ is activated, i.e. $NEL_{\sigma_{in_{1}}, \sigma_{out}}$.state is 1 according to the activating condition of NEL. Consequently, the output neuron $\sigma_{out}$, the post-end of $NEL_{\sigma_{in_{1}}, \sigma_{out}}$, is  activated indicating that computation result of the logic NOT gate is TRUE.
  \item If the inputs are $x_{1}$ = 1, the neurons $\sigma_{in_{1}}$ representing $x_{1}$ is negatively activated, i.e. $\sigma_{in_{1}}$.state=-1. The link $NEL_{\sigma_{in_{1}}, \sigma_{out}}$ of $\sigma_{in_{1}}$ is not activated, i.e. $NEL_{\sigma_{in_{1}}, \sigma_{out}}$.state is 0 according to the activating condition of NEL. Consequently, the output neuron $\sigma_{out}$, the post-end of $NEL_{\sigma_{in_{1}}, \sigma_{out}}$ is  activated indicating that computation result of the logic NOT gate is FALSE.
\end{itemize}

Based on the description of the PLDNN $\prod_{NOT}$ above, it is clear that the PLDNN $\prod_{NOT}$ can correctly emulate the operation of a logic NOT gate, and comparing Figure~\ref{PLDNN NOT} with Figure~\ref{SN P system with astrocyte-like control NOT}, the neural network structure of PLDNN is simpler than SN P systems with astrocyte-like control in emulating the logic gate NOT which has four neurons and more links.

\begin{figure*}[!t]
\centering
\subfloat[PLDNN $\prod_{NOR}$]{\includegraphics[width=0.9\columnwidth]{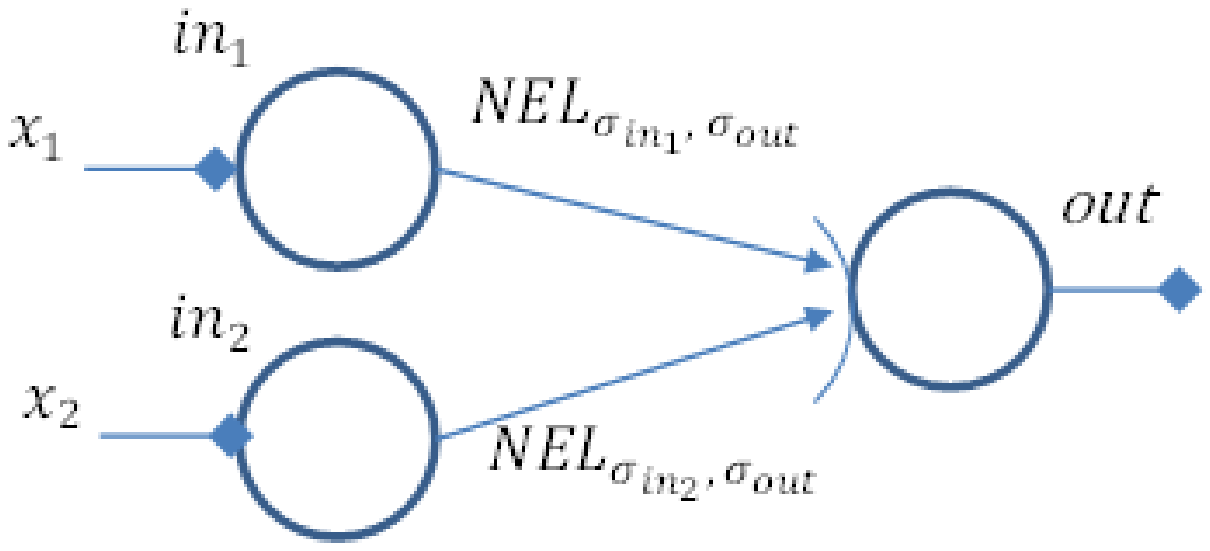}%
\label{PLDNN NOR}}
\hfil
\subfloat[SN P system with astrocyte-like control $\prod_{NOR}$\cite{Tao}]{\includegraphics[width=0.9\columnwidth]{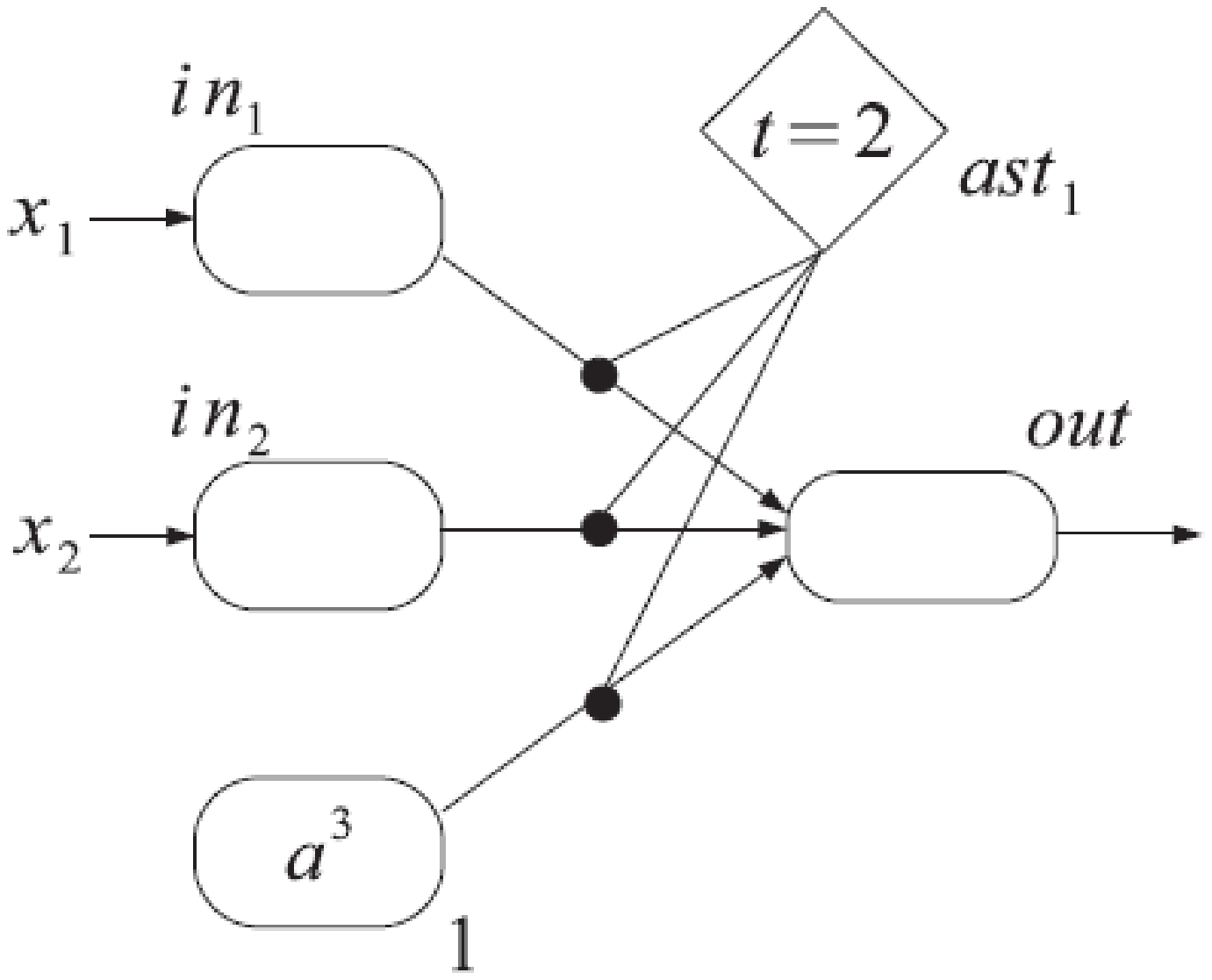}%
\label{SN P system with astrocyte-like control NOR}}
\caption{Simulating a logic NOR gate.}
\label{Simulating a logic NOR gate.}
\end{figure*}

This concludes the proof.
\begin{flushleft}
  \textbf{Theorem 3.4.} The logic NOR gate can be emulated by a simple PLDNN having three neurons and one CEL with two NELs.
\end{flushleft}
\begin{flushleft}
An PLDNN $\prod_{NOR}$ is constructed to emulate the logic gate NOR, shown in Figure~\ref{PLDNN NOR}. The system $\prod_{NOR}$ has two input neurons $\sigma_{in1}$ and $\sigma_{in2}$, and one output neuron $\sigma_{out}$. In the following, all the four cases of inputs to logic NOR gate are considered.
\end{flushleft}
\begin{itemize}
  \item If the inputs are $x_{1}$ = -1 , $x_{2}$ = -1 , the neurons $\sigma_{in_{1}}$ and $\sigma_{in_{2}}$ representing them are negatively activated, i.e. $\sigma_{in_{1}}$.state=-1 and $\sigma_{in_{2}}$.state=-1. The link $NEL_{\sigma_{in_{1}}, \sigma_{out}}$ of $\sigma_{in_{1}}$ is activated, i.e. $NEL_{\sigma_{in_{1}}, \sigma_{out}}$.state is 1 according to the activating condition of NEL. So does the $NEL_{\sigma_{in_{2}}, \sigma_{out}}$, and $NEL_{\sigma_{in_{2}}, \sigma_{out}}$.state is 1. At this moment, the composting exciting link $CEL_{\{\sigma_{in_{1}}, \sigma_{in_{2}}\}, \sigma_{out}}$ is activated according to the activating condition of CEL that all simple ELs should be in the activated state. Consequently, the output neuron $\sigma_{out}$, the post-end of $CEL_{\{\sigma_{in_{1}}, \sigma_{in_{2}}\}, \sigma_{out}}$, is activated indicating that computation result of the logic NOR gate is TRUE.
  \item If the inputs are $x_{1}$ = -1 , $x_{2}$ = 1 , the neuron $\sigma_{in_{1}}$ representing $x_{1}$ is negatively activated and the neuron $\sigma_{in_{2}}$ representing $x_{2}$ is positively activated , i.e. $\sigma_{in_{1}}$.state=-1 and $\sigma_{in_{2}}$.state=1. The link $NEL_{\sigma_{in_{1}}, \sigma_{out}}$ of $\sigma_{in_{1}}$ is activated, i.e. $NEL_{\sigma_{in_{1}}, \sigma_{out}}$.state is 1. $NEL_{\sigma_{in_{2}}, \sigma_{out}}$ is not activated, and $NEL_{\sigma_{in_{2}}, \sigma_{out}}$.state is 0. At this moment, the composting exciting link $CEL_{\{\sigma_{in_{1}}, \sigma_{in_{2}}\}, \sigma_{out}}$ is not activated according to the activating condition of CEL that all simple ELs should be in the activated state. Consequently, the output neuron $\sigma_{out}$, the post-end of $CEL_{\{\sigma_{in_{1}}, \sigma_{in_{2}}\}, \sigma_{out}}$, is not activated indicating that computation result of the logic NOR gate is FALSE.
  \item If the inputs are $x_{1}$ = 1 , $x_{2}$ = -1 , the neuron $\sigma_{in_{1}}$ representing $x_{1}$ is positively activated and the neuron $\sigma_{in_{2}}$ representing $x_{2}$ is negatively activated , i.e. $\sigma_{in_{1}}$.state=1 and $\sigma_{in_{2}}$.state=-1. The computation process of system  NOR is quite similar to the case of inputs being $x_{1}$ = -1 , $x_{2}$ = 1, just exchanging their positions. In this case, the output neuron $\sigma_{out}$, the post-end of $CEL_{\{\sigma_{in_{1}}, \sigma_{in_{2}}\}, \sigma_{out}}$, is not activated indicating that computation result of the logic NOR gate is FALSE.
  \item If the inputs are $x_{1}$ = 1 , $x_{2}$ = 1 , the neurons $\sigma_{in_{1}}$ and $\sigma_{in_{2}}$ representing them are positively activated, i.e. $\sigma_{in_{1}}$.state=1 and $\sigma_{in_{2}}$.state=1. The link $NEL_{\sigma_{in_{1}}, \sigma_{out}}$ of $\sigma_{in_{1}}$ is not activated, i.e. $NEL_{\sigma_{in_{1}}, \sigma_{out}}$.state is 0 according to the activating condition of NEL. So does the $NEL_{\sigma_{in_{2}}, \sigma_{out}}$, and $NEL_{\sigma_{in_{2}}, \sigma_{out}}$.state is 0. At this moment, the composting exciting link $CEL_{\{\sigma_{in_{1}}, \sigma_{in_{2}}\}, \sigma_{out}}$ is not activated according to the activating condition of CEL that all simple ELs should be in the activated state. Eventually, the output neuron $\sigma_{out}$, the post-end of $CEL_{\{\sigma_{in_{1}}, \sigma_{in_{2}}\}, \sigma_{out}}$, is not activated indicating that computation result of the logic NOR gate is FALSE.
\end{itemize}

Based on the description of the PLDNN $\prod_{NOR}$ above, it is clear that the PLDNN $\prod_{NOR}$ can correctly emulate the operation of a logic NOR gate, and comparing Figure~\ref{PLDNN NOR} with Figure~\ref{SN P system with astrocyte-like control NOR}, the neural network structure of PLDNN is simpler than SN P systems with astrocyte-like control in emulating the logic gate NOR which has four neurons and more links.

\begin{figure*}[!t]
\centering
\subfloat[PLDNN $\prod_{XOR}$]{\includegraphics[width=0.9\columnwidth]{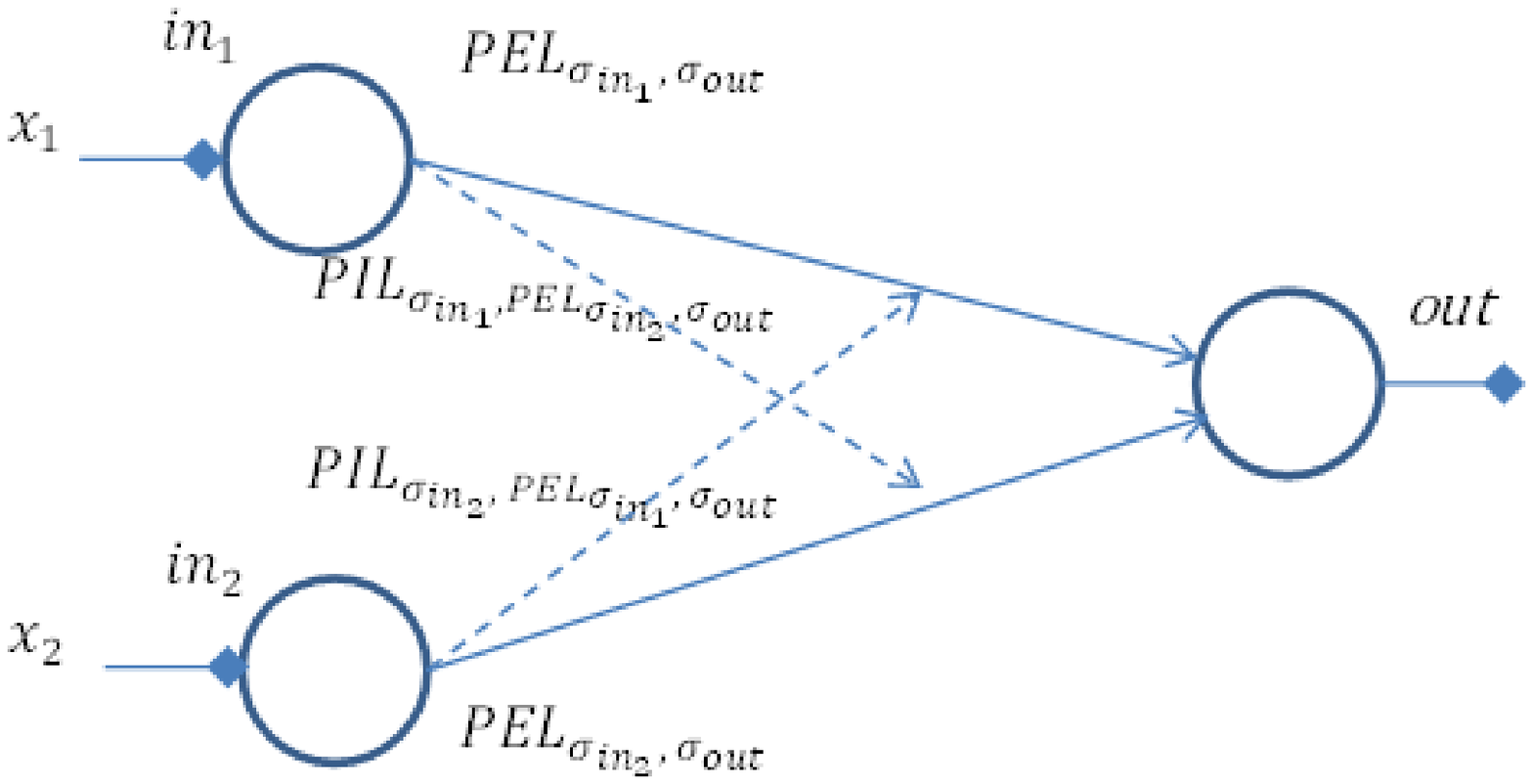}%
\label{PLDNN XOR}}
\hfil
\subfloat[SN P system with astrocyte-like control $\prod_{XOR}$\cite{Tao}]{\includegraphics[width=0.9\columnwidth]{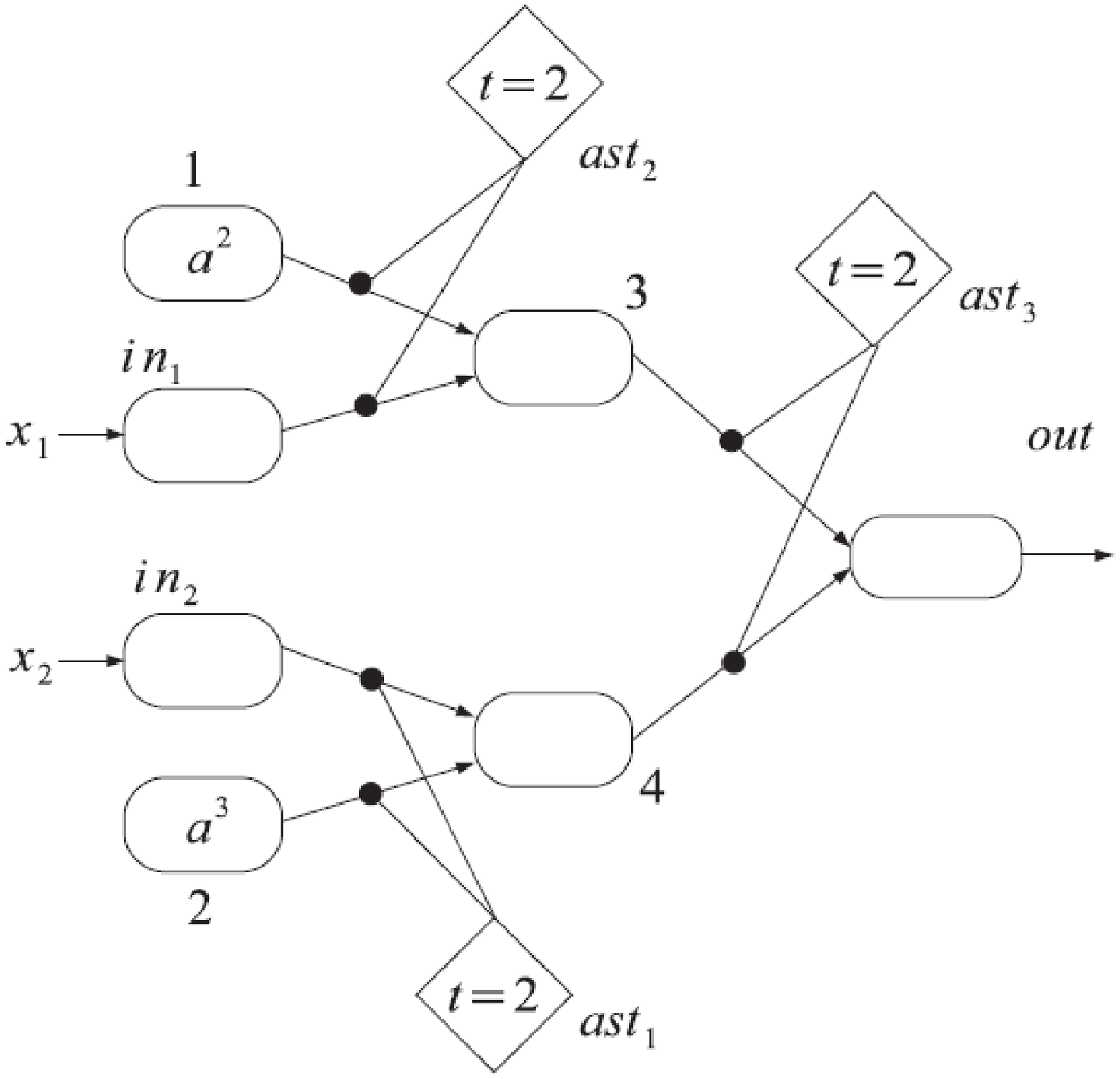}%
\label{SN P system with astrocyte-like control XOR}}
\caption{Simulating a logic XOR gate.}
\label{Simulating a logic XOR gate.}
\end{figure*}

This concludes the proof.
\begin{flushleft}
  \textbf{Theorem 3.5.} The logic XOR gate can be emulated by a simple PLDNN having three neurons, two PELs and two PILs.
\end{flushleft}
\begin{flushleft}
\textbf{Proof.} An PLDNN $\prod_{XOR}$ is constructed to emulate the logic gate XOR, shown in Figure~\ref{PLDNN XOR}. The system $\prod_{XOR}$ has two input neurons $\sigma_{in1}$ and $\sigma_{in2}$, and one output neuron $\sigma_{out}$. In the following, all the four cases of inputs to logic XOR gate are considered.
\end{flushleft}
\begin{itemize}
  \item If the inputs are $x_{1}$ = -1 , $x_{2}$ = -1 , the neurons $\sigma_{in_{1}}$ and $\sigma_{in_{2}}$ representing them are negatively activated, i.e. $\sigma_{in_{1}}$.state=-1 and $\sigma_{in_{2}}$.state=-1. The link $PEL_{\sigma_{in_{1}}, \sigma_{out}}$ of $\sigma_{in_{1}}$ is not activated, i.e. $PEL_{\sigma_{in_{1}}, \sigma_{out}}$.state is 0 according to the activating condition of PEL. So does the $PEL_{\sigma_{in_{2}}, \sigma_{out}}$, and $PEL_{\sigma_{in_{2}}, \sigma_{out}}$.state is 0. Consequently, the output neuron $\sigma_{out}$, the post-end of $PEL_{\sigma_{in_{1}}, \sigma_{out}}$ and $PEL_{\sigma_{in_{2}}, \sigma_{out}}$, is not activated indicating that computation result of the logic XOR gate is FALSE.
  \item If the inputs are $x_{1}$ = -1 , $x_{2}$ = 1, the neuron $\sigma_{in_{1}}$ representing $x_{1}$ is negatively activated and the neuron $\sigma_{in_{2}}$ representing $x_{2}$ is positively activated , i.e. $\sigma_{in_{1}}$.state=-1 and $\sigma_{in_{2}}$.state=1. The link $PEL_{\sigma_{in_{1}}, \sigma_{out}}$ of $\sigma_{in_{1}}$ is not activated, i.e. $PEL_{\sigma_{in_{1}}, \sigma_{out}}$.state is 0. $PEL_{\sigma_{in_{2}}, \sigma_{out}}$ is activated, and $PEL_{\sigma_{in_{2}}, \sigma_{out}}$.state is 1. Consequently, the output neuron $\sigma_{out}$ is activated by $PEL_{\sigma_{in_{2}}, \sigma_{out}}$, indicating that computation result of the logic XOR gate is TRUE.
  \item If the inputs are $x_{1}$ = 1 , $x_{2}$ = -1 , the neuron $\sigma_{in_{1}}$ representing $x_{1}$ is positively activated and the neuron $\sigma_{in_{2}}$ representing $x_{2}$ is negatively activated , i.e. $\sigma_{in_{1}}$.state=1 and $\sigma_{in_{2}}$.state=-1. The computation process of system  OR is quite similar to the case of inputs being $x_{1}$ = -1 , $x_{2}$ = 1, just exchanging their positions. In this case, the output neuron $\sigma_{out}$ is activated by $PEL_{\sigma_{in_{2}}, \sigma_{out}}$ indicating that computation result of the logic XOR gate is TRUE.
  \item If the inputs are $x_{1}$ = 1 , $x_{2}$ = 1 , the neurons $\sigma_{in_{1}}$ and $\sigma_{in_{2}}$ representing them are positively activated, i.e. $\sigma_{in_{1}}$.state=1 and $\sigma_{in_{2}}$.state=1. The link $PEL_{\sigma_{in_{1}}, \sigma_{out}}$ of $\sigma_{in_{1}}$ is activated, i.e. $PEL_{\sigma_{in_{1}}, \sigma_{out}}$.state is 1 according to the activating condition of PEL. So does the $PEL_{\sigma_{in_{2}}, \sigma_{out}}$, and $PEL_{\sigma_{in_{2}}, \sigma_{out}}$.state is 1. The link $PIL_{\sigma_{in_{1}}, \sigma_{out}}$ of $\sigma_{in_{1}}$ is activated, i.e. $PIL_{\sigma_{in_{1}}, \sigma_{out}}$.state is 1 according to the activating condition of PIL. The inhibiting effect of $PIL_{\sigma_{in_{1}}, \sigma_{out}}$ works in the activated state, inhibiting $PEL_{\sigma_{in_{2}}, \sigma_{out}}$ from activating $\sigma_{out}$. So does the $PIL_{\sigma_{in_{2}}, \sigma_{out}}$, and $PIL_{\sigma_{in_{2}}, \sigma_{out}}$.state is 1. Eventually, the output neuron $\sigma_{out}$ is not activated indicating that computation result of the logic XOR gate is FALSE.
\end{itemize}

Based on the description of the PLDNN $\prod_{XOR}$ above, it is clear that the PLDNN $\prod_{XOR}$ can correctly emulate the operation of a logic XOR gate, and comparing Figure~\ref{PLDNN XOR} with Figure~\ref{SN P system with astrocyte-like control XOR}, the neural network structure of PLDNN is simpler than SN P systems with astrocyte-like control in emulating the logic gate XOR which has seven neurons and more links.

\begin{figure*}[!t]
\centering
\subfloat[PLDNN $\prod_{NAND}$]{\includegraphics[width=0.9\columnwidth]{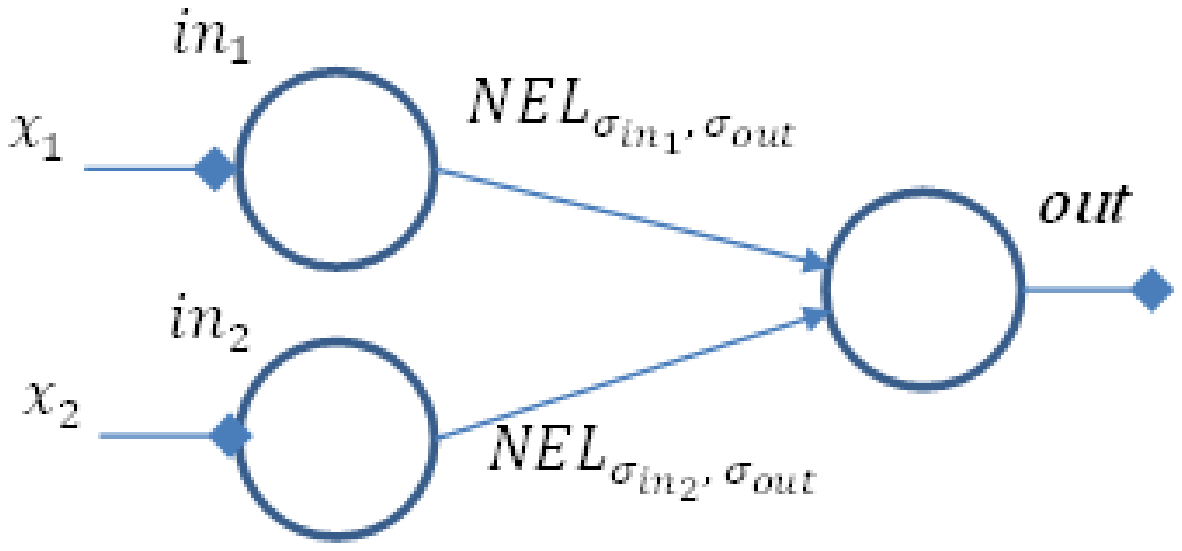}%
\label{PLDNN NAND}}
\hfil
\subfloat[SN P system with astrocyte-like control $\prod_{NAND}$\cite{Tao}]{\includegraphics[width=0.9\columnwidth]{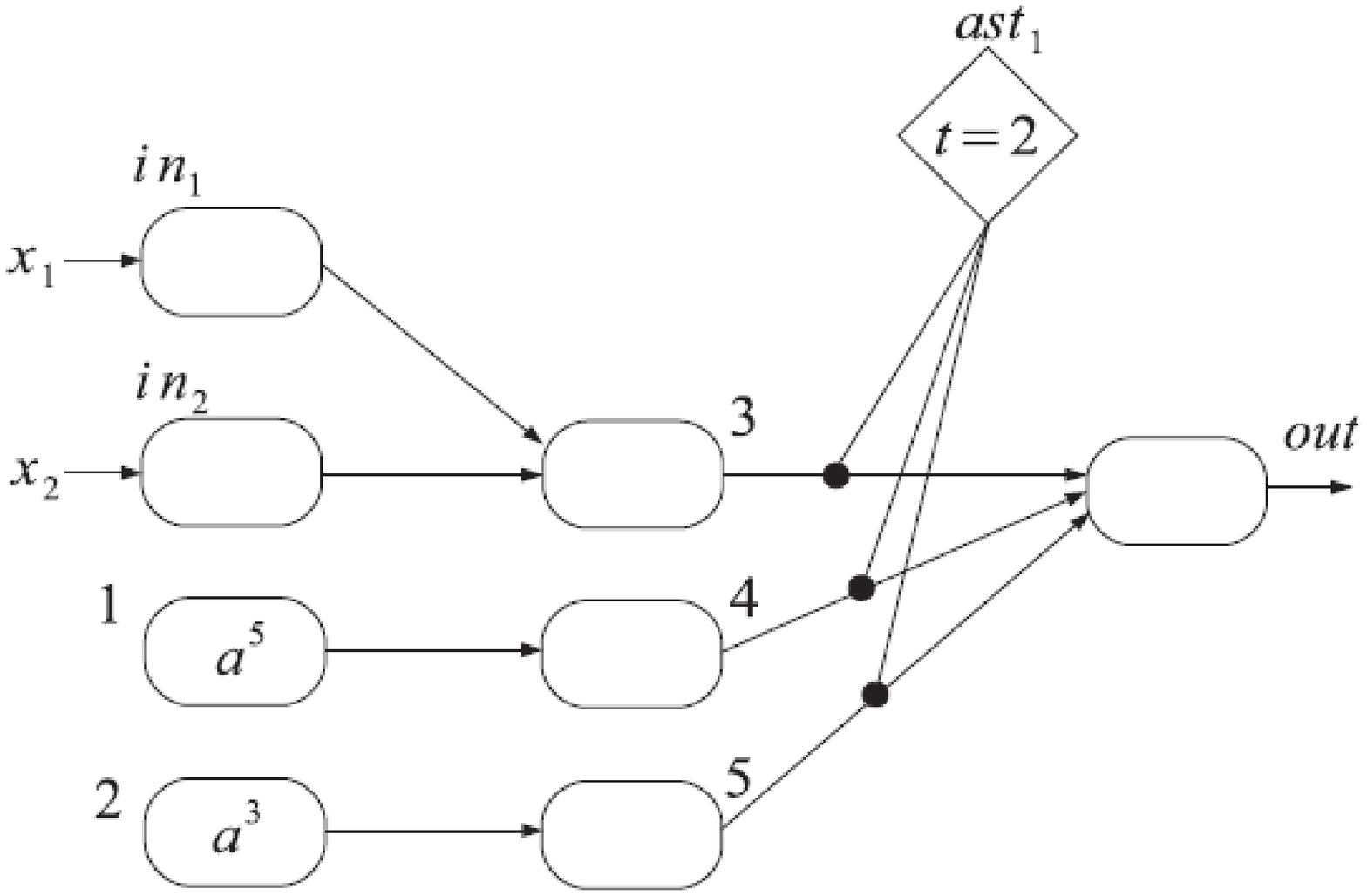}%
\label{SN P system with astrocyte-like control NAND}}
\caption{Simulating a logic NAND gate.}
\label{Simulating a logic NAND gate.}
\end{figure*}

This concludes the proof.
\begin{flushleft}
  \textbf{Theorem 3.6.} The logic NAND gate can be emulated by a simple PLDNN having three neurons and two NELs.
\end{flushleft}
\begin{flushleft}
\textbf{Proof.} An PLDNN $\prod_{NAND}$ is constructed to emulate the logic gate NAND, shown in Figure~\ref{PLDNN NAND}. The system $\prod_{NAND}$ has two input neurons $\sigma_{in1}$ and $\sigma_{in2}$, and one output neuron $\sigma_{out}$. In the following, all the four cases of inputs to logic NAND gate are considered.
\end{flushleft}
\begin{itemize}
  \item If the inputs are $x_{1}$ = -1 , $x_{2}$ = -1 , the neurons $\sigma_{in_{1}}$ and $\sigma_{in_{2}}$ representing them are negatively activated, i.e. $\sigma_{in_{1}}$.state=-1 and $\sigma_{in_{2}}$.state=-1. The link $NEL_{\sigma_{in_{1}}, \sigma_{out}}$ of $\sigma_{in_{1}}$ is activated, i.e. $NEL_{\sigma_{in_{1}}, \sigma_{out}}$.state is 1 according to the activating condition of NEL. So does the $NEL_{\sigma_{in_{2}}, \sigma_{out}}$, and $NEL_{\sigma_{in_{2}}, \sigma_{out}}$.state is 1. Consequently, the output neuron $\sigma_{out}$, the post-end of $CEL_{\{\sigma_{in_{1}}, \sigma_{in_{2}}\}, \sigma_{out}}$, is activated indicating that computation result of the logic NAND gate is TRUE.
  \item If the inputs are $x_{1}$ = -1 , $x_{2}$ = 1 , the neuron $\sigma_{in_{1}}$ representing $x_{1}$ is negatively activated and the neuron $\sigma_{in_{2}}$ representing $x_{2}$ is positively activated , i.e. $\sigma_{in_{1}}$.state=-1 and $\sigma_{in_{2}}$.state=1. The link $NEL_{\sigma_{in_{1}}, \sigma_{out}}$ of $\sigma_{in_{1}}$ is activated, i.e. $NEL_{\sigma_{in_{1}}, \sigma_{out}}$.state is 1. $NEL_{\sigma_{in_{2}}, \sigma_{out}}$ is not activated, and $NEL_{\sigma_{in_{2}}, \sigma_{out}}$.state is 0. Consequently, the output neuron $\sigma_{out}$ is activated indicating that computation result of the logic NAND gate is TRUE.
  \item If the inputs are $x_{1}$ = 1 , $x_{2}$ = -1 , the neuron $\sigma_{in_{1}}$ representing $x_{1}$ is positively activated and the neuron $\sigma_{in_{2}}$ representing $x_{2}$ is negatively activated , i.e. $\sigma_{in_{1}}$.state=1 and $\sigma_{in_{2}}$.state=-1. The computation process of system  NOR is quite similar to the case of inputs being $x_{1}$ = -1 , $x_{2}$ = 1, just exchanging their positions. In this case, the output neuron $\sigma_{out}$ is activated indicating that computation result of the logic NAND gate is TRUE.
  \item If the inputs are $x_{1}$ = 1 , $x_{2}$ = 1 , the neurons $\sigma_{in_{1}}$ and $\sigma_{in_{2}}$ representing them are positively activated, i.e. $\sigma_{in_{1}}$.state=1 and $\sigma_{in_{2}}$.state=1. The link $NEL_{\sigma_{in_{1}}, \sigma_{out}}$ of $\sigma_{in_{1}}$ is not activated, i.e. $NEL_{\sigma_{in_{1}}, \sigma_{out}}$.state is 0 according to the activating condition of NEL. So does the $NEL_{\sigma_{in_{2}}, \sigma_{out}}$, and $NEL_{\sigma_{in_{2}}, \sigma_{out}}$.state is 0. Eventually, the output neuron $\sigma_{out}$, the post-end of both $NEL_{\{\sigma_{in_{1}}, \sigma_{out}}$ and $NEL_{\{\sigma_{in_{2}}, \sigma_{out}}$, is not activated indicating that computation result of the logic NAND gate is FALSE.
\end{itemize}

Based on the description of the PLDNN $\prod_{NAND}$ above, it is clear that the PLDNN $\prod_{NAND}$ can correctly emulate the operation of a logic NAND gate, and comparing Figure~\ref{PLDNN NAND} with Figure~\ref{SN P system with astrocyte-like control NAND}, the neural network structure of PLDNN is simpler than SN P systems with astrocyte-like control in emulating the logic gate NAND which has eight neurons and more links.

This concludes the proof.
\section{An Example of Representing Logical Expression in Neural-Like Style}
\begin{figure*}[!t]
\centering
\subfloat[PLDNN $\prod_{NAND}$]{\includegraphics[width=0.9\columnwidth]{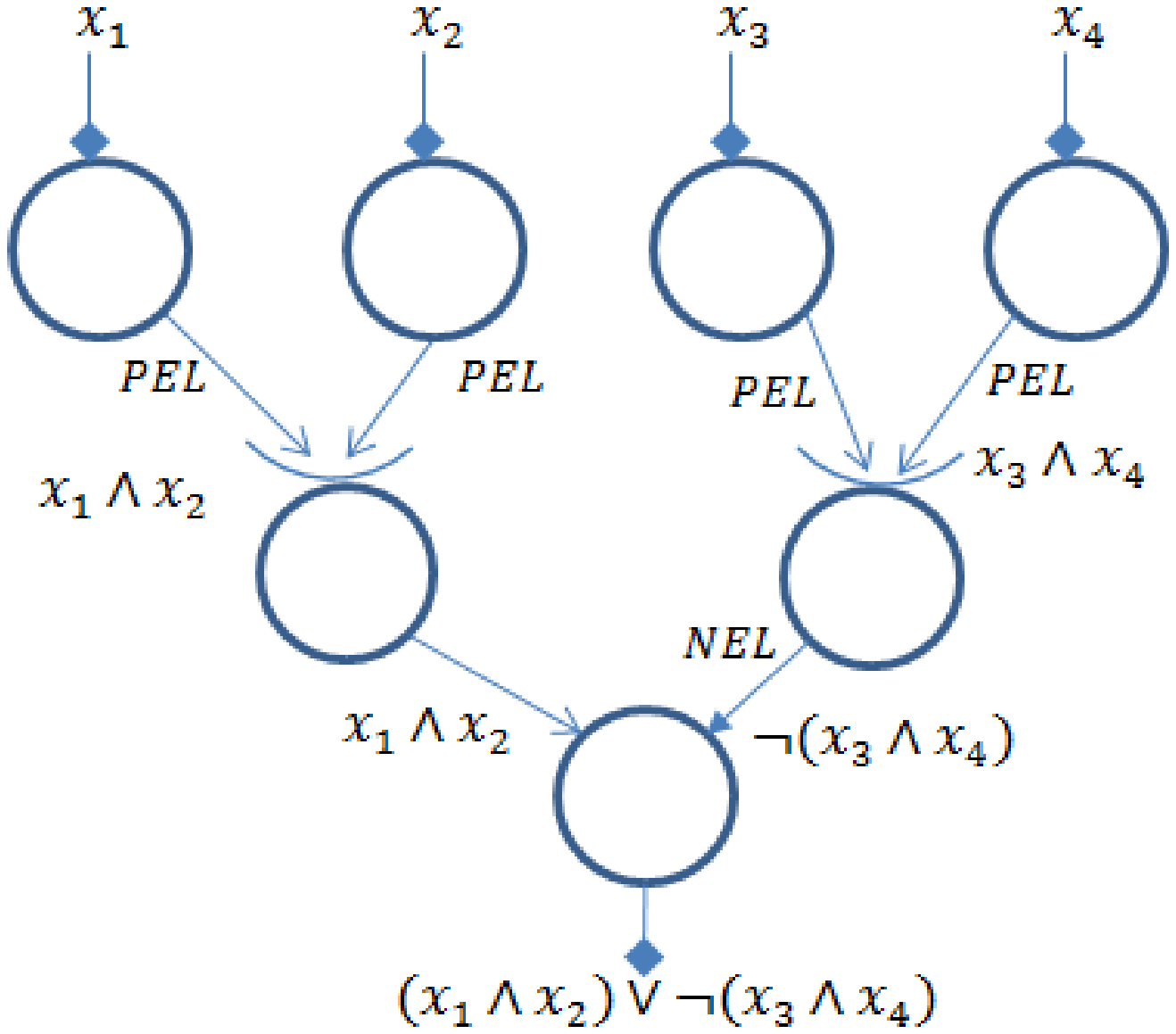}%
\label{PLDNN expressionp}}
\hfil
\subfloat[SN P system with astrocyte-like control $\prod_{NAND}$\cite{Tao}]{\includegraphics[width=0.9\columnwidth]{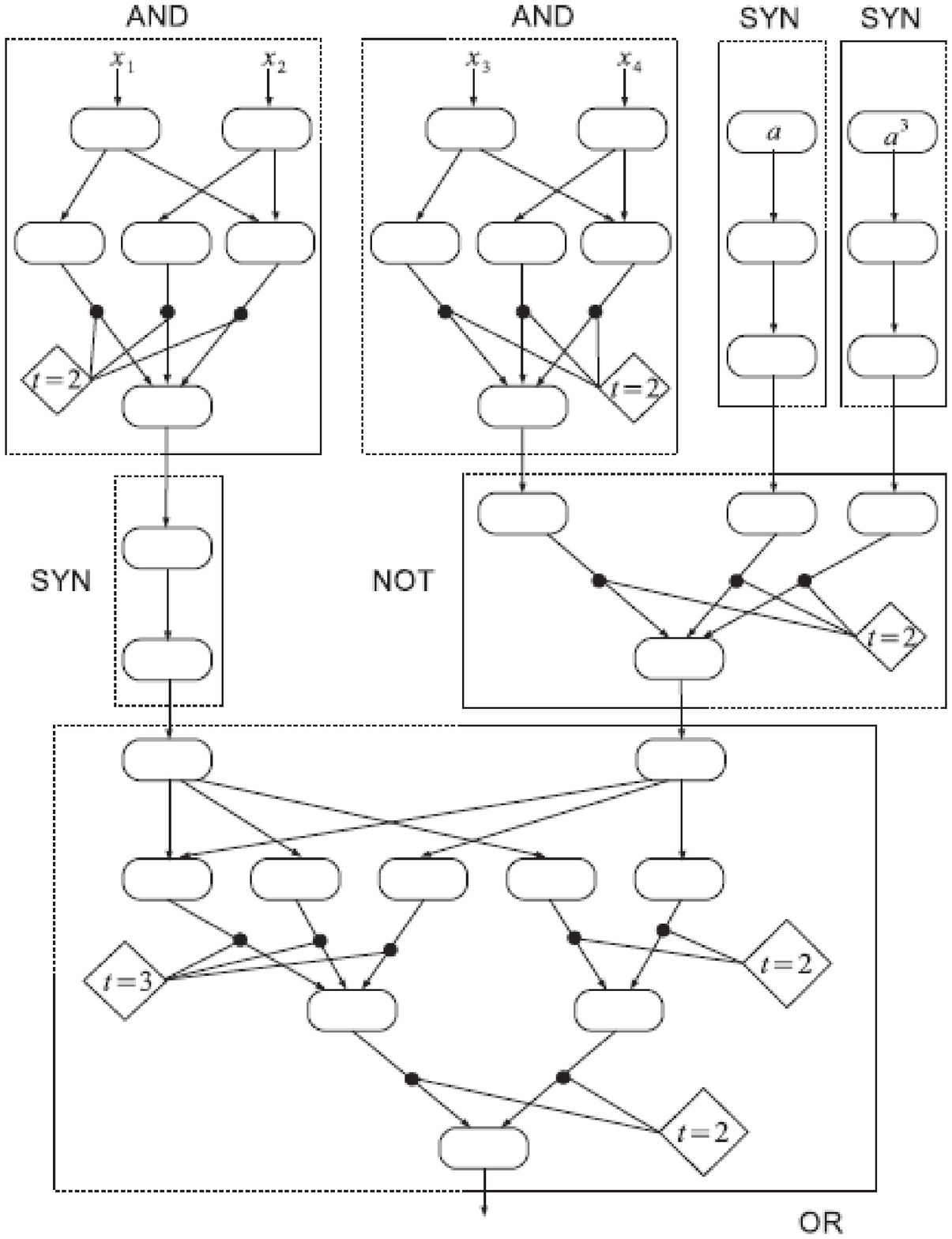}%
\label{SN P system with astrocyte-like control expressions}}
\caption{Simulating the logical expression $(x_{1}\bigwedge x_{2})\bigvee \neg(x_{3}\bigwedge x_{4})$.}
\label{Simulating the logical expression.}
\end{figure*}

With the neural-like logic gates designed in section~\ref{Simulating logic gates}, an example of representing logical expressions is developed by connecting neural-like logic gates emulated by PLDNN proposed. The logic gates constructed in section~\ref{Simulating logic gates} are applied to represent the logical expression $(x_{1}\bigwedge x_{2})\bigvee \neg(x_{3}\bigwedge x_{4})$ in \cite{Ionescu}. It can be easily checked that the PLDNN constructed in Figure~\ref{Simulating the logical expression.} can represent this logical relation. The details are omitted here. Through the comparison of the two systems as shown in Figure~\ref{Simulating the logical expression.}, it is obvious that PLDNN has simpler neural network structure than SN P system with asytrocyte-like control in representing this logical relation. PLDNN has less neurons and links.

\begin{figure}[!h]
\begin{center}
\includegraphics[width=0.9\columnwidth]{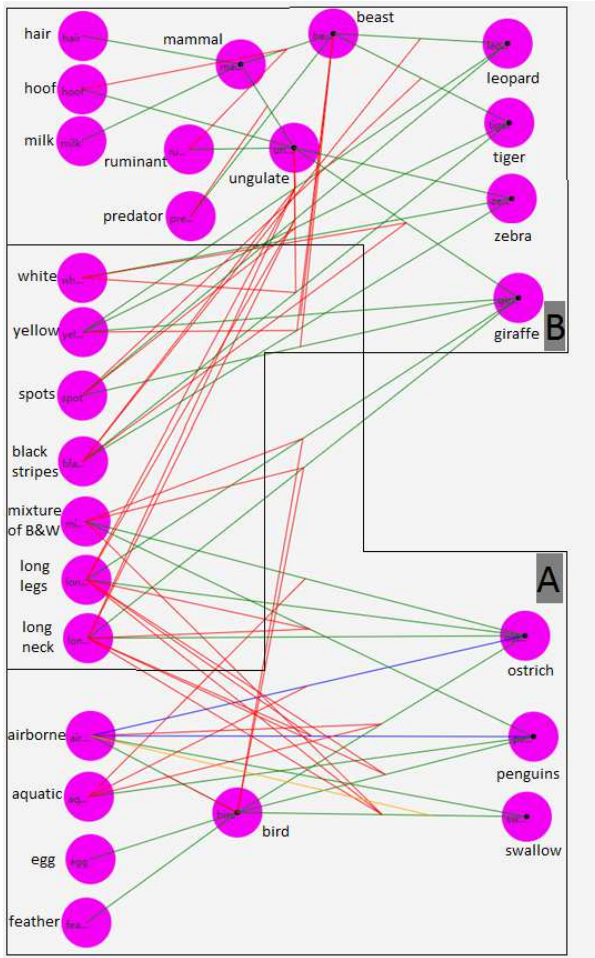}
\caption{Demo: Network structure of PLDNN to represent logical relations of animals and their attributes}
\label{structure}
\end{center}
\end{figure}

In the following, an intuitive and piratical demo of PLDNN to learn the rule library containing 14 logical relations about animals(see appendix~\ref{Supplements}). These relations are usually used as example in AI related books and materials like Haykin¡¯s Neural Networks and Learning Machine \cite{Simon}. The network structure of PLDNN  after learning is shown in Figure~\ref{structure}. If the rule library is developed by the heave-weight logical components of SN P system with asytrocyte-like control, it can be imaged that the developing process is hard and its network structure is more complex than PLDNN since it needs more neurons and links.

The network structure of PLDNN forms a knowledge graph. The knowledge graph is represented through the interconnection structure of the neural network. Colors are used to distinguish links instead of the legends of links in the paper for the flexibility of programming. The network structure represents logical relations. Green indicates PEL, blue indicates NEL, red indicates PIL, and orange indicates NIL. From Figure~\ref{structure}, we can see that the network structure of PLDNN is created according to the data. If things have no relations with each other, the neurons representing them have no links between each other. PLDNN uses the network structure to represent and store logical relations. The PLDNN in Figure~\ref{structure} can be manually verified that whether PLDNN memories the relations based on commonsense. For example, when PLDNN perceives a data (yellow and black strips), the neuron Y representing yellow will excite the following neurons L,T and G(leopard, tiger and giraffe) connected by Y's PELs. The neuron BS representing black strips will excite the following neurons T and Z (tiger and zebra) connected by BS's PELs. BS's PILs inhibits Y's PELs to excite the neurons L and G, and Y's PIL inhibits BS's PEL to excite the neuron Z. After these interactions between neurons through links, PLDNN reasons that the animal is tiger given the data (yellow and black strips).

\section{Conclusion}
In this research, we proposed and established PLDNN to emulate logic AND, OR, NOT, NOR, XOR and NAND gates. Compared with SN P systems with astrocyte-like control, PLDNN has a simpler structure while SN P systems has a complex structure with excessive numbers of neurons and links to supply so as to emulate logic gates. The difference can be intuitively observed in the example of representing a logical expression. Beneficial for the specialization in representing logical relations, less neurons and links are used in PLDNN. The design pattern of PLDNN makes the responsibilities between the neurons and link not overlap, i.e. neurons are specialized only for representing things while links between neurons are specialized only for representing logical relations between things. No extra neurons are needed to deal with representing relations.

As demonstrated in the example of representing a logical expression, this work provides a novel and easy way of constructing ¡°neural-like¡±logic gates and representing complex logical expressions by combination. It may point a different potential direction to represent logical expression by using neural network theoretically. As a new candidate model in neural networks, there are still further research in the application and optimization. For example, in the application of PLDNN, how to get features in a domain like medicine and represent them by neurons of PLDNN is the next work.

\section{}
\label{Supplements}
In the following, it is a simple rule library as example which contains 14 logical relations. The PLDNN in Figure~\ref{structure} memorizes them and forms a knowledge graph through the interconnection structure of the neural network.
\begin{enumerate}
  \item If an animal has hair,  then it is mammal
  \item If an animal produces milk,  then it is mammal
  \item If a mammal is predator,  then it is beast
  \item If a mammal has hoof,  then it is ungulate
  \item If a mammal is ruminant, then it is ungulate
  \item If an animal has feather, produces egg,  then it is bird
  \item If an animal airborne, then it is bird
  \item If a beast is yellow and spots,  then it is leopard
  \item If a beast is yellow and black strips,  then it is tiger
  \item If an ungulate has long neck, long leg, yellow and spots, then it is giraffe
  \item If an ungulate is white and black strips,  then it is zebra
  \item If a bird cannot airborne, has long neck, long legs, and is mixture of black and white,  then it is ostrich
  \item If a bird cannot airborne, can aquatic, and is mixture of black and white, then it is penguin
  \item If a bird can airborne,  then it is swallow
\end{enumerate}

These relations are usually used as example in AI-related book and materials like Haykin's Neural Networks and Learning Machines.


\ifCLASSOPTIONcaptionsoff
  \newpage
\fi



%

\end{document}